\documentclass{article}

\usepackage{PRIMEarxiv}

\usepackage[utf8]{inputenc}
\usepackage[T1]{fontenc}
\usepackage{hyperref}
\usepackage{url}
\usepackage{booktabs}
\usepackage{amsfonts}
\usepackage{amssymb}
\usepackage{amsmath}
\usepackage{nicefrac}
\usepackage{microtype}
\usepackage{graphicx}
\usepackage{multirow}
\usepackage{array}
\usepackage{siunitx}
\usepackage{xcolor}
\usepackage{listings}
\usepackage{tikz}
\usepackage{pgfplots}
\usepackage{pgfplotstable}
\pgfplotsset{compat=1.18}

\definecolor{codegreen}{rgb}{0,0.6,0}
\definecolor{codegray}{rgb}{0.5,0.5,0.5}
\definecolor{codepurple}{rgb}{0.58,0,0.82}
\definecolor{backcolour}{rgb}{0.97,0.97,0.97}

\title{
    \includegraphics[width=\textwidth]{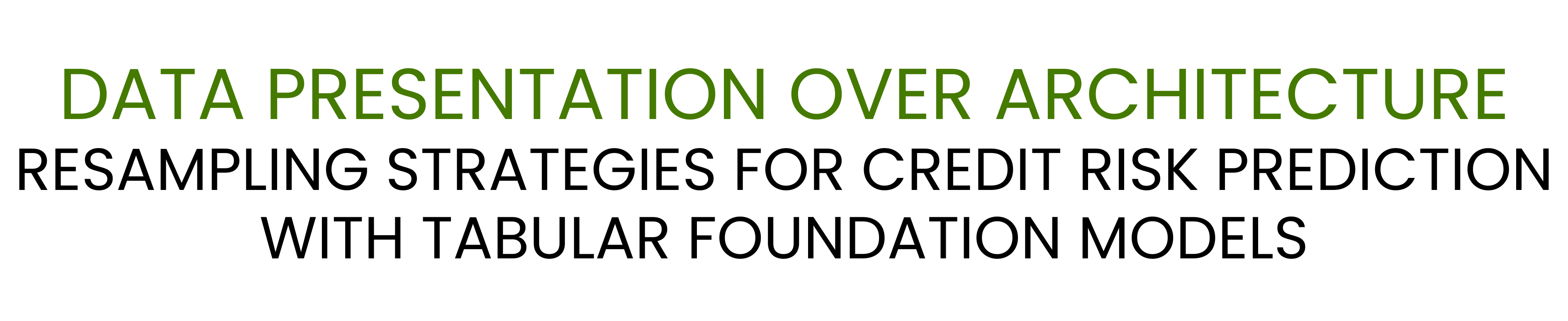}
}

\author{
  Aditya Tanna , Mitul Solanki, Mohamed Bouadi, Nassim Bouarour, \\
  Pratinav Seth, Vinay Kumar Sankarapu \\
  \affiliation{Lexsi Labs}\\
  \texttt{aditya.tanna@lexsi.ai}
}

\setabstract{%
Credit default prediction is a tabular learning problem with severe class imbalance, heterogeneous features, and tight latency budgets. Tabular Foundation Models (TFMs) approach this problem through in-context learning, which makes their predictions sensitive to how the context window is built. We benchmark four classical models and five TFMs on the Home Credit and Lending Club datasets, varying the context-construction strategy (seven options) and the context size (1K to 50K). On both datasets, the choice of context strategy explains more variance in AUC-ROC than the choice of TFM family: balanced and hybrid sampling add 3 to 4 AUC points over uniform sampling, and the gap exceeds the spread between TFMs. With a balanced context of 5K to 10K examples, the strongest TFMs reach the AUC of classical baselines trained on the full data, while also recovering meaningful default-class recall that default-threshold GBDTs do not. We frame this as evidence that context construction, rather than architecture choice, is the primary deployment lever for TFMs in imbalanced credit-risk settings.
}

\setkeywords{Credit risk prediction, Tabular foundation models, Class imbalance, Context construction, Financial data systems, Resampling strategies}

\runningtitle{Data Presentation Over Architecture}

\begin{document}
\maketitle

\section{Introduction}
\label{sec:introduction}

Credit default prediction underpins lending decisions across the financial industry. Despite decades of research, production credit systems face persistent challenges: severe class imbalance, informative missingness, and predictive signals that emerge from complex feature interactions rather than from individual variables. Gradient-boosted decision trees (GBDTs) such as XGBoost, LightGBM, and CatBoost dominate tabular benchmarks and production deployments~\cite{grinsztajn2022tree}, benefiting from inductive biases well-suited to heterogeneous tabular data. Yet even well-tuned GBDTs can collapse to majority-class prediction under severe imbalance when deployed without threshold adjustment, yielding near-zero minority recall despite high aggregate accuracy.

An alternative paradigm, \emph{Tabular Foundation Models} (TFMs), reframes prediction as in-context learning (ICL). Rather than training per-dataset, TFMs (e.g., TabPFN~\cite{hollmann2023tabpfn}, TabICL~\cite{qu2025tabicl}, OrionMSP~\cite{bouadi2025orionmsp}, and OrionBix~\cite{bouadi2025orionbix}) condition on a \emph{context window} of labeled examples and make predictions via a single forward pass analogous to few-shot learning in language models. Crucially, the composition of the context is a design choice independent from model architecture, making TFMs sensitive to \emph{data presentation}. This raises a relevant question for financial systems: \emph{Does data composition matter as much as model architecture for performance?} If so, the focus should shift from model selection to \emph{intelligent data curation}, which is fundamentally a data-management problem.

Prior work establishes that tree-based models outperform deep neural networks on typical tabular data~\cite{grinsztajn2022tree}, with class imbalance addressed in classical ML through resampling and cost-sensitive learning~\cite{chawla2002smote,he2009learning}. Recent benchmarks evaluate TFMs across curated tabular suites~\cite{gardner2024benchmarking} but ignore the interaction between \emph{context composition} and severe class imbalance in financial settings. Our work here is to treat context construction as a performance lever for financial TFMs, linking deployment to data management. We propose seven context-construction strategies and evaluate them on a controlled benchmark across two large credit risk datasets (Home Credit and Lending Club), using five TFMs based on the TabTune~\cite{tanna2025tabtune} framework, and comparing the TFMs to four classical baselines.

\paragraph{Contributions.}
This paper makes the following contributions:
\begin{itemize}
\setlength{\itemsep}{2pt}
\setlength{\parsep}{0pt}
\item \textbf{Data composition dominates architecture.} Balanced and hybrid resampling yield 3--4\% AUC gains, exceeding the typical gap between TFMs regardless of dataset or model family.
\item \textbf{TFMs are data-efficient.} TFMs match classical models trained on the full dataset using only 5K--10K balanced samples, with no gradient optimization and a single forward pass at inference.
\item \textbf{Resampling resolves the zero-recall trap.} TFMs with balanced context achieve MCC $\approx$ 0.2 and default-class F1 of 0.24--0.31, producing meaningful default detection. Classical baselines at default thresholds predict the majority class almost exclusively (recall $\approx$ 0\%).
\item \textbf{A reframing of credit-risk modeling as a data-systems problem.} We argue that context construction is a design axis for ICL pipelines on a par with model selection: representative, class-aware, budget-efficient subsets of financial data become first-class system artifacts.
\end{itemize}

\begin{figure*}[t]
  \centering
  \includegraphics[width=0.95\textwidth]{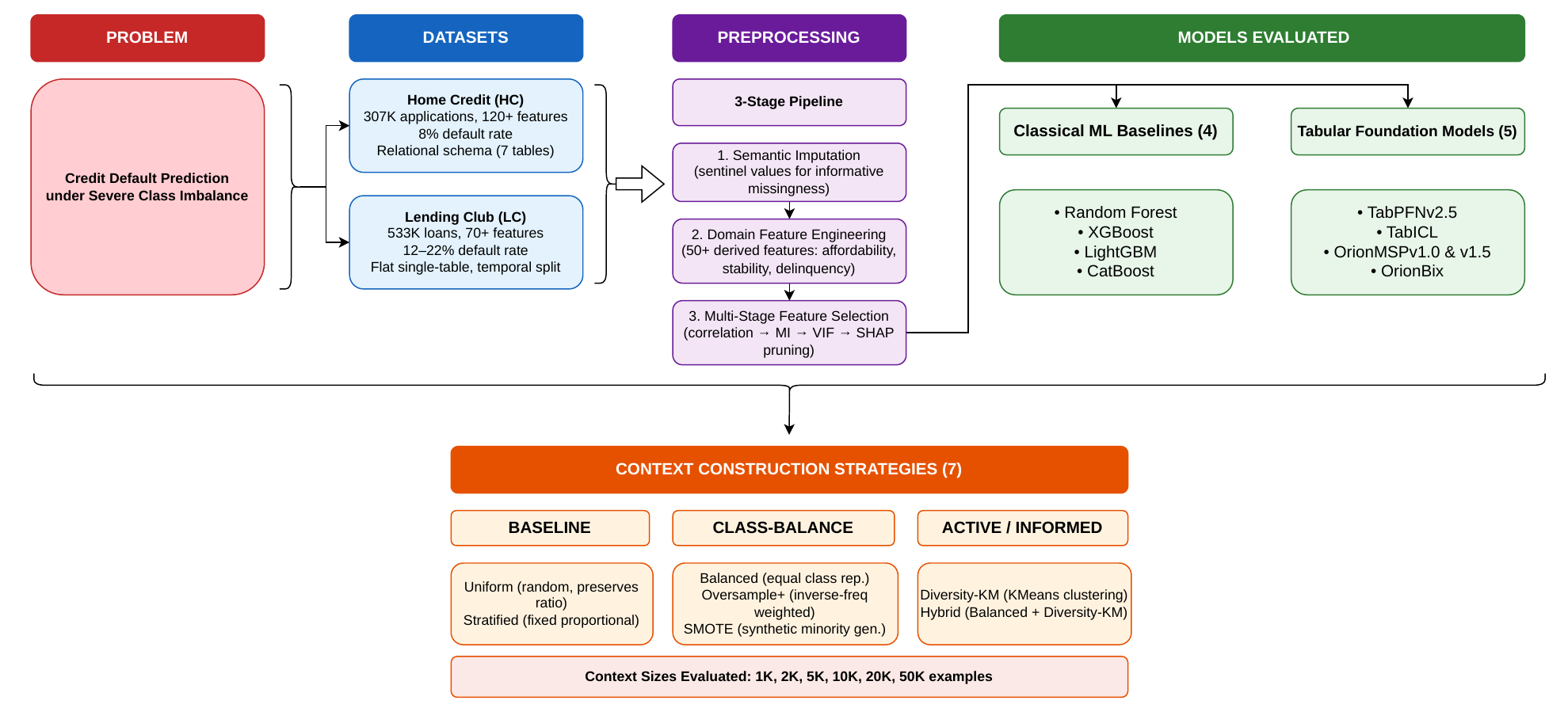}
  \caption{Overview of the study. Two credit-risk datasets (Home Credit, Lending Club) are passed through a three-stage preprocessing pipeline (semantic imputation, domain feature engineering, multi-stage feature selection). Four classical baselines and five Tabular Foundation Models are evaluated under seven context-construction strategies grouped into three families (Baseline, Class-balance, Active/Informed), at six context sizes from 1K to 50K.}
  \label{fig:overview}
\end{figure*}

\section{Related Work}
\label{subsec:related-work}

\textbf{Tabular foundation models and in-context learning.} TabPFN~\cite{hollmann2023tabpfn} introduced the idea of a transformer pre-trained on synthetic tabular tasks that performs Bayesian inference via a single forward pass on a small context. Subsequent work has scaled this paradigm to larger and more heterogeneous tabular settings (TabICL~\cite{qu2025tabicl}, OrionMSP~\cite{bouadi2025orionmsp}, OrionBix~\cite{bouadi2025orionbix}). Surveys~\cite{van2024tabular,borisov2022deep} catalogue this line of work and the broader space of deep tabular models. The TabTune library~\cite{tanna2025tabtune} provides a unified interface for inference and fine-tuning across TFM variants, which we use as the experimental backbone of this study.

\textbf{Tabular benchmarks and TFM evaluation.} Grinsztajn et al.~\cite{grinsztajn2022tree} report that tree-based models retain an edge on medium-sized tabular data, and Gardner et al.~\cite{gardner2024benchmarking} systematically benchmark TFMs on curated suites. These works focus on aggregate accuracy across balanced or mildly imbalanced datasets; they do not examine how the context window is constructed under severe class imbalance, which is the regime that matters for credit-risk deployment.

\textbf{Class imbalance and credit risk.} Imbalanced classification has a long history in tabular ML, with SMOTE~\cite{chawla2002smote}, ADASYN~\cite{he2008adasyn}, and cost-sensitive learning~\cite{he2009learning} as standard tools, and \texttt{imbalanced-learn}~\cite{lemaitre2017imblearn} as a common implementation. In credit scoring specifically, Baesens et al.~\cite{baesens2003benchmarking} and Thomas et al.~\cite{thomas2017credit} document the dominance of statistical and tree-based models, and the operational importance of minority-class detection. Our work treats these resampling techniques as \emph{context-construction strategies} rather than training-time data augmentations, which is a reframing made possible by the ICL setting.

\textbf{In-context example selection.} A parallel literature in NLP studies how the choice of in-context examples affects LLM performance~\cite{liu2024incontext,rubin2022learning}, typically via similarity-based retrieval. Our findings echo this line of work on the basic point that which examples appear in the context matters, but the tabular regime adds a class-balance axis that is largely absent from NLP-ICL.

\section{Problem Statement and Study Design}
\label{sec:problem}

We consider supervised binary credit-risk prediction where the positive class (default) is substantially underrepresented. In tabular in-context learning, a TFM does not consume the full training set at inference time, but a bounded context of labeled examples. Once that budget is fixed, performance depends on two choices: the model architecture and the strategy used to construct the context. We therefore ask: \textit{Given an imbalanced credit dataset and a fixed context budget, how much does predictive performance depend on context-construction strategy relative to model architecture, and are these effects consistent across TFM families?}

This question has a direct data-systems interpretation. If architecture is the dominant factor, effort should focus on model search and tuning. If context construction dominates, data selection becomes a first-class system design problem: how to build representative, class-aware, budget-efficient subsets of financial data for inference. Our hypothesis is that, under severe imbalance, minority underrepresentation within a bounded context is a key bottleneck, making class-aware context construction a high-leverage deployment decision. If this holds consistently across TFM families, it reframes data curation as a generalizable engineering recommendation rather than a model-specific tuning artifact.

To test this, we conduct a controlled study on two large-scale credit benchmarks, Home Credit and Lending Club, comparing four classical baselines and five TFMs across seven context-construction strategies and context sizes from 1K to 50K. In addition to ROC-AUC, we report recall and Matthews Correlation Coefficient (MCC) to assess minority-class detection. Figure~\ref{fig:overview} summarizes the full pipeline.

\section{Experimental Setup}
\label{sec:setup}

\subsection{Datasets}
We evaluate on two widely-used credit risk benchmarks:

\textbf{Home Credit Default Risk}~\cite{homecredit2018} ({\sc HC}) comprises $\sim$307K loan applications with 120+ features derived from a relational schema of seven tables (bureau records, previous applications, payment histories). The default rate is $\sim$8\%, with heavy missingness ($>$50\% in many fields). Predictive power emerges primarily from feature interactions between income, credit exposure, and repayment behavior rather than from individual variables.

\textbf{Lending Club}~\cite{lendingclub2020} ({\sc LC}) contains $\sim$533K resolved consumer loans with 70+ borrower- and loan-level features. The default rate is $\sim$12--22\% depending on filtering. We use temporal splitting (train $\leq$ June 2019; test: H2 2019) to prevent leakage.

Both datasets undergo a three-stage preprocessing pipeline.
(1)~\emph{Semantic imputation:} sentinel values ($-1$) encode informative missingness rather than replacing it with statistical estimates. Absent bureau scores signal thin-file customers, missing car age signals no vehicle ownership, and refused-loan dates distinguish ``never approved'' from ``not yet applied.''
(2)~\emph{Feature engineering:} 50+ derived features capture affordability ratios (credit-to-income, annuity-to-income, repayment rate), employment stability proxies, delinquency severity aggregates across active bureau loans, and behavioral flags from installment payment history.
(3)~\emph{Multi-stage feature selection:} correlation filtering $\rightarrow$ mutual information ranking $\rightarrow$ VIF analysis $\rightarrow$ SHAP-based importance pruning. The pipeline is structured but deliberately \emph{not} competition-optimized, preserving comparability across model families.

\subsection{Models}
\textbf{Classical baselines.} Random Forest~\cite{breiman2001random}, XGBoost~\cite{chen2016xgboost}, LightGBM~\cite{ke2017lightgbm}, and CatBoost~\cite{prokhorenkova2018catboost}, trained on the full processed dataset with default hyperparameters and \emph{no} class-weighting adjustments or threshold tuning. This controlled evaluation matches TFM conditions: both paradigms are assessed at their out-of-the-box operating point, ensuring that observed differences reflect data composition rather than post-hoc calibration.

\textbf{Tabular Foundation Models.} Five TFMs evaluated through TabTune~\cite{tanna2025tabtune}, a unified experimentation framework for tabular foundation models: TabPFN~\cite{hollmann2023tabpfn}, TabICL~\cite{qu2025tabicl}, OrionMSP v1.0 and v1.5 ~\cite{bouadi2025orionmsp}, and OrionBix~\cite{bouadi2025orionbix}. Each conditions predictions on a context window of labeled examples at sizes $\{1024, 2048, 5000, 10000, 20000, 50000\}$.

\subsection{Context-Construction Strategies}
Since TFMs operate on a finite context window, the composition of this window is a design choice. We evaluate seven strategies grouped into three families:

\noindent\textbf{Baseline.} \textit{Uniform} (random sampling, preserves the original class ratio), \textit{Stratified} (fixed proportional sampling, reduces run-to-run variance without rebalancing).

\noindent\textbf{Class-balance.} \textit{Balanced} (equal class representation, $m_c \approx m/K$ samples per class $c$), \textit{Oversample+} (inverse-frequency weighted sampling with boost multiplier and minimum minority count), \textit{SMOTE}~\cite{chawla2002smote} (synthetic minority generation via nearest-neighbour interpolation).

\noindent\textbf{Active/Informed.} \textit{Diversity-KM} (MiniBatch KMeans clustering; one representative per cluster, maximizing feature-space coverage), \textit{Hybrid} (fraction $\rho$ from Balanced, remainder from Diversity-KM; combines minority signal amplification with representational breadth).

\subsection{Evaluation}
ROC-AUC is the primary metric, following credit risk convention. We additionally report Matthews Correlation Coefficient (MCC) and default-class recall to assess minority detection, metrics that expose the ``zero-recall trap'' which accuracy alone obscures.

\section{Results}
\label{sec:results}

\subsection{Default-Threshold Failure on Imbalanced Data}
\label{subsec:zero-recall}
Table~\ref{tab:baselines} reports classical baseline performance under default operating points. All four models reach 87--92\% accuracy on both datasets, but on {\sc HC} three of them assign zero positive predictions: every applicant is classified as non-default. This is a textbook instance of the accuracy paradox under severe class imbalance~\cite{provost2001robust, he2009learning, japkowicz2002class}: a classifier optimized for 0/1 loss at a 0.5 threshold collapses to the majority class when the minority rate falls below roughly 10\%. We use \emph{zero-recall regime} as shorthand for this operating-point failure throughout the paper, with the caveat that it is a manifestation of a known phenomenon rather than a new one. What is worth flagging is that even gradient-boosted trees with categorical handling exhibit it on {\sc HC} without re-weighting or threshold tuning, and that, as we show in §\ref{subsec:resolving}, balanced context construction resolves it for TFMs in the same way \texttt{scale\_pos\_weight} resolves it for GBDTs.

\begin{table}[t]
\caption{Classical baseline results. Best AUC per dataset in \textbf{bold}. Recall is default-class recall. Models evaluated at default threshold without class-weighting.}
\label{tab:baselines}
\small
\centering
\begin{tabular}{llcccc}
\toprule
& \textbf{Model} & \textbf{AUC} & \textbf{Acc} & \textbf{Recall} & \textbf{MCC} \\
\midrule
\multirow{4}{*}{\rotatebox[origin=c]{90}{\scriptsize Home Cr.}}
& Random Forest & \textbf{0.739} & 0.920 & 0.5\% & 0.055 \\
& XGBoost       & 0.719          & 0.919 & 0.0\% & 0.000 \\
& LightGBM      & 0.713          & 0.919 & 0.0\% & 0.000 \\
& CatBoost      & 0.638          & 0.919 & 0.0\% & 0.000 \\
\midrule
\multirow{4}{*}{\rotatebox[origin=c]{90}{\scriptsize Lend.~Cl.}}
& XGBoost       & \textbf{0.718} & 0.869 & 7.2\% & 0.117 \\
& Random Forest & 0.703          & 0.871 & 3.2\% & 0.069 \\
& CatBoost      & 0.655          & 0.870 & 3.2\% & 0.060 \\
& LightGBM      & 0.650          & 0.875 & 0.0\% & 0.000 \\
\bottomrule
\end{tabular}
\end{table}

\subsection{Strategy Matters More Than Model}

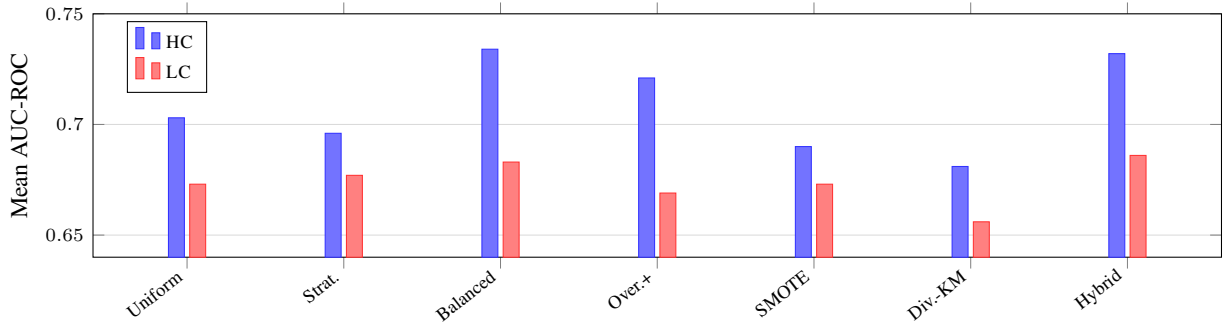
\begin{figure}[t]
\centering
\begin{tikzpicture}
\begin{axis}[
  width=\columnwidth,
  height=4.8cm,
  ylabel={Mean AUC-ROC},
  ymin=0.640, ymax=0.750,
  xtick={1,2,3,4,5,6,7},
  xticklabels={Uniform, Strat., Balanced, Over.+, SMOTE, Div.-KM, Hybrid},
  xticklabel style={rotate=38, anchor=east, font=\scriptsize},
  ybar=2pt,
  bar width=6pt,
  enlarge x limits=0.10,
  legend pos=north west,
  legend style={font=\scriptsize, cells={anchor=west}},
  ymajorgrids=true,
  grid style={line width=0.3pt, draw=gray!30},
  tick label style={font=\scriptsize},
  label style={font=\small},
]
\addplot[fill=blue!55, draw=blue!80] coordinates {
  (1,0.703)(2,0.696)(3,0.734)(4,0.721)(5,0.690)(6,0.681)(7,0.732)
};
\addplot[fill=red!50, draw=red!80] coordinates {
  (1,0.673)(2,0.677)(3,0.683)(4,0.669)(5,0.673)(6,0.656)(7,0.686)
};
\legend{{\sc HC}, {\sc LC}}
\end{axis}
\end{tikzpicture}
\caption{Mean AUC-ROC by context-strategy, averaged across all TFMs and context sizes. Balanced leads on {\sc HC}; Hybrid leads on {\sc LC}. The best--worst strategy gap ($\sim$0.03--0.05) exceeds the typical inter-TFM gap within any single strategy.}
\label{fig:strategy}
\end{figure}

Figure~\ref{fig:strategy} and Table~\ref{tab:strategy} present mean AUC-ROC for each context-construction strategy, averaged across all five TFMs and six context sizes (30 experiments per strategy per dataset). The central finding: \textbf{the choice of resampling strategy produces larger AUC differences than the choice of TFM architecture.} Balanced achieves the highest mean AUC on {\sc HC} (0.734), and Hybrid leads on {\sc LC} (0.686). In both cases, the gap between the best and worst strategy ($\sim$0.03--0.05) exceeds the inter-TFM gap within any single strategy.

Pairwise win-rate analysis reinforces this: Balanced wins $\geq$70\% of head-to-head matchups against most competitors on {\sc HC} and ranks first in $\sim$50\% of individual experiments; Hybrid achieves comparable dominance on {\sc LC}. SMOTE and Diversity-KM show high variance and model-dependent effects, making them unreliable as general-purpose strategies. The consistent pattern across both datasets is that class-awareness in context construction is more valuable than feature-space coverage alone.

\begin{figure}[t]
\centering
\includegraphics[width=0.85\columnwidth]{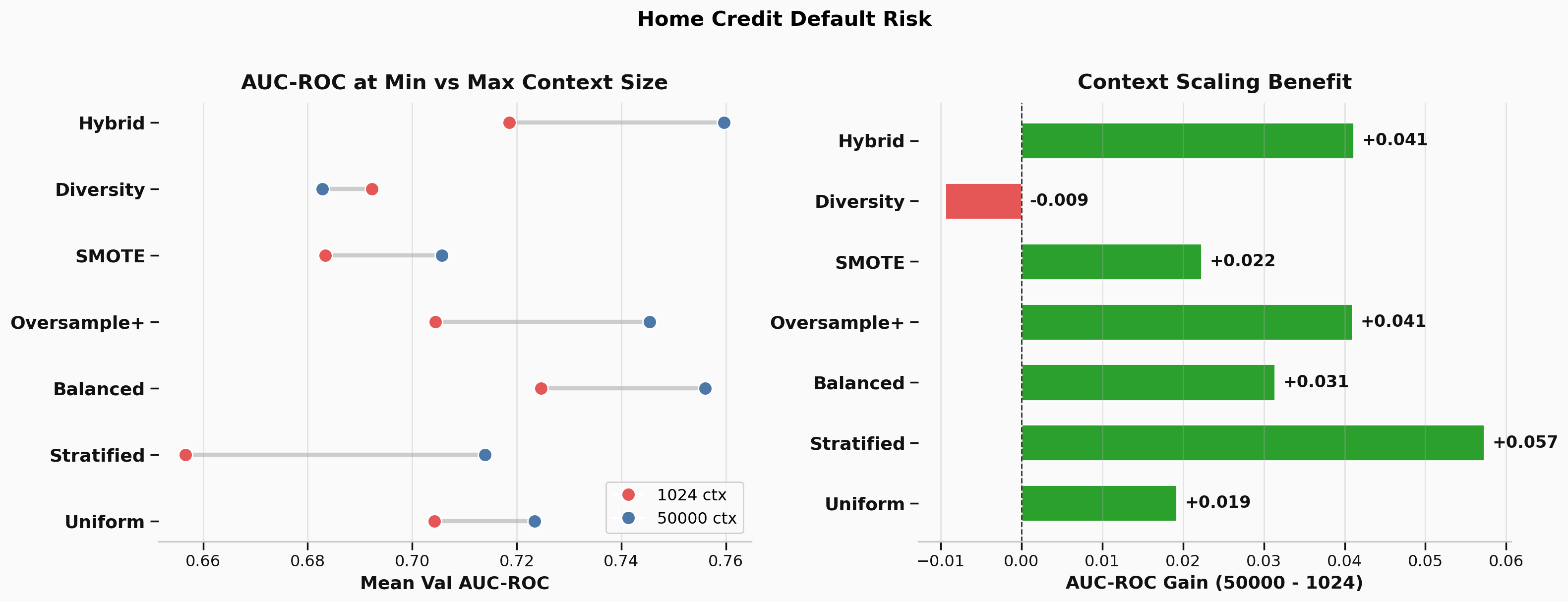}\\[6pt]
\includegraphics[width=0.85\columnwidth]{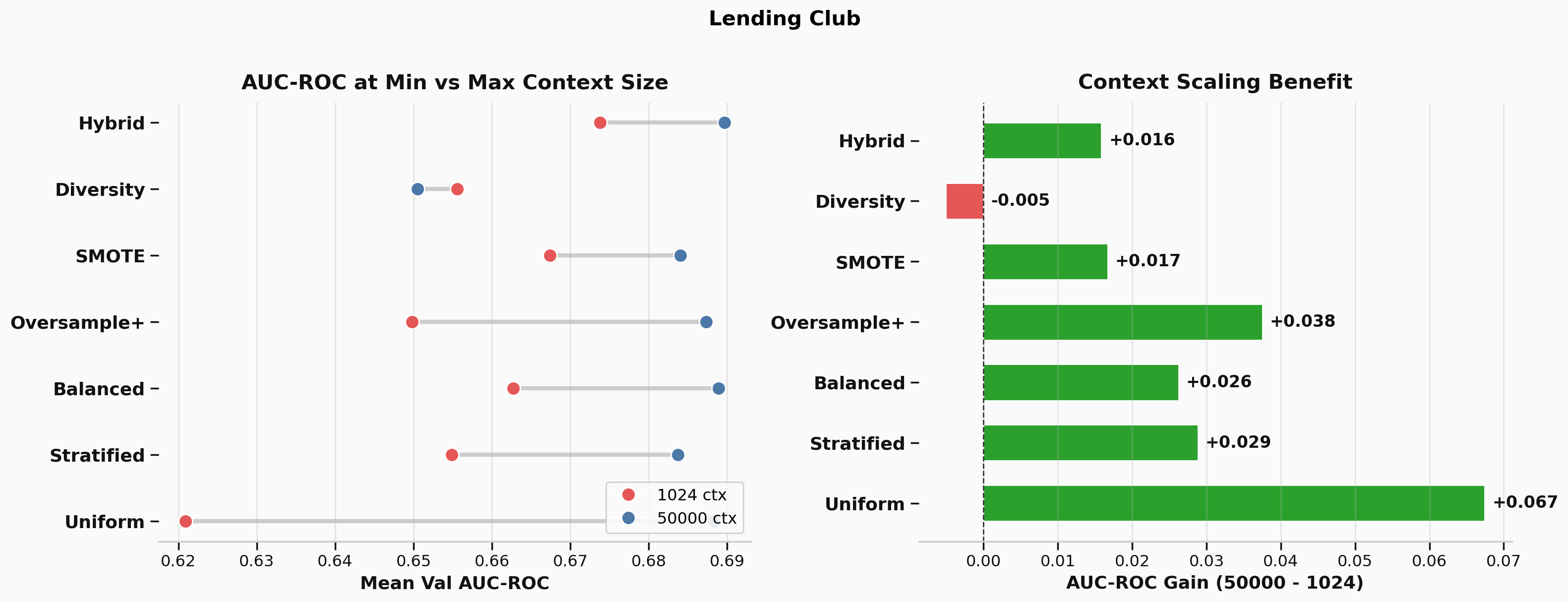}
\caption{AUC-ROC at minimum (1K) vs.\ maximum (50K) context size (left) and absolute scaling gain (right) per resampling strategy. \textbf{(a) {\sc HC}:} Balanced and Hybrid start high and scale further (+0.034, +0.045); Diversity-KM degrades with scale ($-$0.016), confirming class-unawareness introduces noise. \textbf{(b) {\sc LC}:} Uniform shows the largest gain (+0.069) but from a lower base; Balanced and Hybrid maintain the highest AUC at both extremes.}
\label{fig:context_scaling}
\end{figure}

\subsection{Data Efficiency: The 5K--10K Crossover}

TFM performance broadly improves with context size: the two strongest models (TabPFN, TabICL) show consistent gains up to 50K, while OrionMSP and OrionBix plateau above 10K. Critically, TabPFN and TabICL \textbf{surpass the best classical baseline} (Random Forest, 0.739) between 5K and 10K context samples, representing a \textbf{25--50$\times$ data reduction}: matching models trained on 246K samples using $\sim$5--10K in-context examples, with no gradient training, no hyperparameter search, and a single forward pass.

On {\sc LC}, the crossover against XGBoost (0.718) is not reached within 50K context for most TFMs (best at 50K: TabPFN, 0.703). We attribute this to the dataset's flat single-table structure and higher base default rate ($\sim$15\%), which favor gradient-based models that can exploit dense, interaction-free feature signals across the full 533K training set.

\begin{table}[t]
\caption{Mean AUC-ROC by context-strategy, averaged across all TFMs and context sizes. Best per dataset in \textbf{bold}. Uniform acts as a baseline. See Figure~\ref{fig:strategy} for visual comparison.}
\label{tab:strategy}
\small
\centering
\begin{tabular}{lcc}
\toprule
\textbf{Strategy} & \textbf{{\sc HC}} & \textbf{{\sc LC}} \\
\midrule
Uniform        & 0.703 & 0.673 \\
Stratified     & 0.696 & 0.677 \\
\textbf{Balanced}  & \textbf{0.734} & 0.683 \\
Oversample+    & 0.721 & 0.669 \\
SMOTE          & 0.690 & 0.673 \\
Diversity-KM   & 0.681 & 0.656 \\
\textbf{Hybrid}    & 0.732 & \textbf{0.686} \\
\bottomrule
\end{tabular}
\end{table}

\subsection{Resolving the Zero-Recall Trap}
\label{subsec:resolving}

Table~\ref{tab:mcc} shows the operational impact of balanced context construction. Classical baselines achieve MCC near zero, they provide essentially no information about default risk beyond predicting the majority class. At 50K balanced context, all five TFMs achieve MCC between 0.19 and 0.26, with default-class F1 scores of 0.24--0.31 and balanced accuracy of 0.65--0.71.

\textit{This improvement is not architectural}: TFMs under Uniform sampling also suffer from the zero-recall trap, achieving MCC $\approx$ 0 on {\sc HC}. The gain comes \emph{entirely} from presenting the model with a class-balanced context window. TFM architectures have sufficient capacity to discriminate defaulters when given adequate minority exposure; the model is not the bottleneck, the data presentation is.

\textbf{Why balanced context helps.} This is consistent with a straightforward view of TFM in-context learning: the context window acts as an implicit prior over the prediction distribution. A context that is 92\% non-default (as on Home Credit under Uniform sampling) pulls the model toward predicting non-default, the same operating-point failure that traps default-threshold GBDTs. Balancing the context corrects that prior without retraining the model or tuning the threshold, which is why a 5K--10K balanced window recovers minority recall on every TFM family we tested.

\begin{table}[t]
\caption{MCC and default-class F1 at 50K balanced context across both datasets. Classical baselines evaluated on full training data; TFMs evaluated under balanced context construction. Best per column per dataset in \textbf{bold}.}
\label{tab:mcc}
\small
\centering
\begin{tabular}{lcccc}
\toprule
\textbf{Model} & \textbf{AUC} & \textbf{MCC} & \textbf{Def.~F1} & \textbf{Bal.~Acc} \\
\midrule
\multicolumn{5}{l}{\textit{\textbf{{\sc HC}} --- Classical baselines}} \\
Random Forest    & \textbf{0.739} & 0.055 & 0.010 & \textbf{0.502} \\
XGBoost          & 0.719 & 0.000 & 0.000 & 0.500 \\
CatBoost         & 0.638 & 0.000 & 0.000 & 0.500 \\
LightGBM         & 0.713 & 0.000 & 0.000 & 0.500 \\
\midrule
\multicolumn{5}{l}{\textit{\textbf{{\sc HC}} --- TFMs at 50K balanced context}} \\
TabPFN           & \textbf{0.786} & \textbf{0.258} & 0.302          & \textbf{0.711} \\
TabICL           & 0.771          & 0.245          & 0.299          & 0.691 \\
OrionMSP         & 0.738          & 0.213          & 0.278          & 0.662 \\
OrionBix         & 0.733          & 0.192          & 0.242          & 0.672 \\
OrionMSPv1.5     & 0.700          & 0.197          & \textbf{0.310} & 0.647 \\
\midrule
\multicolumn{5}{l}{\textit{\textbf{{\sc LC}} --- Classical baselines}} \\
XGBoost          & \textbf{0.718} & \textbf{0.117} & \textbf{0.121} & \textbf{0.527} \\
Random Forest    & 0.703 & 0.069 & 0.058 & 0.511 \\
CatBoost         & 0.655 & 0.060 & 0.058 & 0.511 \\
LightGBM         & 0.650 & 0.000 & 0.000 & 0.500 \\
\midrule
\multicolumn{5}{l}{\textit{\textbf{{\sc LC}} --- TFMs at 50K balanced context}} \\
TabICL           & \textbf{0.713} & 0.202          & 0.313          & 0.651 \\
TabPFN           & 0.705          & \textbf{0.203} & \textbf{0.317} & \textbf{0.650} \\
OrionMSP         & 0.690          & 0.180          & 0.301          & 0.633 \\
OrionMSPv1.5     & 0.690          & 0.190          & 0.306          & 0.642 \\
OrionBix         & 0.647          & 0.128          & 0.268          & 0.597 \\
\bottomrule
\end{tabular}
\end{table}

\subsection{Cross-Dataset Consistency}

The strategy-dominates-architecture finding holds across both datasets despite their structural differences (relational vs.\ flat, 8\% vs.\ 12--22\% default rate, 246K vs.\ 533K samples, random vs.\ temporal split). On {\sc LC}, the Balanced--Hybrid gap narrows and Hybrid edges ahead, possibly because the higher base default rate reduces imbalance severity, shifting the primary bottleneck from minority exposure toward representational breadth. Diversity-KM consistently underperforms on both datasets, confirming that feature-space coverage without class awareness introduces noise rather than signal at these imbalance levels.

\section{Discussion}
\label{sec:discussion-related}

\subsection{Design Decisions}
\label{subsec:design-decisions}

\textbf{Why default-threshold baselines.} We deliberately evaluate classical baselines at their default operating point, without class-weighting or threshold tuning. The same choice is applied to TFMs, which receive no calibration either. This is the cleanest like-for-like comparison: any performance difference reflects what the model receives at training/inference time, not a downstream calibration step. Adding \texttt{scale\_pos\_weight} would predictably restore default recall for GBDTs, but it would conflate composition effects with post-hoc calibration.

\textbf{Why these seven strategies.} The seven strategies are chosen to span the three obvious families a practitioner would consider: do nothing special (Uniform, Stratified), rebalance the classes (Balanced, Oversample+, SMOTE), or pick examples that cover the feature space (Diversity-KM, Hybrid). We are not claiming these are the best strategies; we are claiming that the gap between them is large enough to dominate the gap between TFM architectures, which is the load-bearing claim of the paper.

\textbf{Why two datasets.} {\sc HC} and {\sc LC} are deliberately different: relational vs.\ flat, $\sim$8\% vs.\ 12--22\% default rate, $\sim$246K vs.\ 533K training samples, random vs.\ temporal split. Consistent qualitative findings across both reduce the risk that the result is an artifact of one schema or one preprocessing pipeline.

\subsection{Implications for Financial Data Systems}

Our results reframe credit risk modeling as a \emph{data curation problem}: rather than investing in model architecture search, financial data pipelines should prioritize constructing representative, class-balanced training subsets. Data selection, quality, and composition become first-class system requirements, positioning context construction as a new design axis for in-context learning pipelines.

\textbf{Operational considerations.} TFMs with balanced context require no retraining when data distributions shift, and the 5K--10K crossover implies reduced storage and compute for model refresh cycles. However, TFM inference latency (seconds vs.\ milliseconds per sample) limits real-time applicability; GBDTs remain pragmatic when full labeled data and latency budgets are available.

\subsection{Limitations}
\label{subsec:limitations}

Our preprocessing is structured but not competition-optimized; top {\sc HC} solutions reach $\sim$0.80 AUC via ensembling across all seven source tables, which we deliberately exclude to isolate data composition effects. Classical baselines use default thresholds without class-weighting to match TFM conditions; applying \texttt{scale\_pos\_weight} would improve recall but conflate composition effects with post-hoc calibration. The seven context-construction strategies are heuristic; learned context selection (e.g., similarity- or uncertainty-based retrieval) is an obvious extension we do not study here. Finally, our evaluation is limited to two consumer-credit datasets; corporate credit, SME lending, and country-level macroeconomic stress scenarios may show different sensitivity patterns.

\textbf{Future directions.} Fine-tuning TFMs on domain-specific labeled data~\cite{tanna2026finetuning} and replacing heuristic resampling with learned context selection are natural next steps, with the strategies identified here serving as strong baselines for both.

\section{Conclusion}
\label{sec:conclusion}

We benchmarked classical ML and five Tabular Foundation Models on two large-scale credit risk datasets across seven context-construction strategies and six context sizes. The central finding is narrower than ``data over architecture'': \emph{among TFMs, the strategy used to fill the context window explains more variance in AUC than the choice of TFM family}, and this holds across both datasets despite their different schemas, default rates, and split protocols. For financial data systems, this motivates treating training data curation as a first-class engineering concern alongside model selection, and positions context construction as a new design axis for ICL pipelines.

\bibliographystyle{unsrt}
\bibliography{references}

\end{document}